\pdfoutput=1

\documentclass[11pt]{article}

\usepackage[final]{acl}

\usepackage{times}
\usepackage{latexsym}
\usepackage{hyperref}       
\usepackage{url}            
\usepackage{booktabs}       
\usepackage{amsfonts}       
\usepackage{nicefrac}       
\usepackage{xcolor}         
\usepackage[most]{tcolorbox}
\usepackage{times}
\usepackage{latexsym}
\usepackage{amsmath}
\usepackage{makecell}
\usepackage{multirow}

\usepackage[T1]{fontenc}

\usepackage[utf8]{inputenc}

\usepackage{microtype}

\usepackage{inconsolata}

\usepackage{graphicx}

\title{Palette of Language Models: A Solver for Controlled Text Generation}

\author{Zhe Yang,Yi Huang\thanks{Corresponding authors}, Yaqin Chen, Xiaoting Wu, Junlan Feng and Chao Deng\\
  JIUTIAN Team, China Mobile Research Institute\\
  \texttt{\{yangzhe,huangyi,chenyaqin,wuxiaoting,fengjunlan,dengchao\}@chinamobile.com} \\}
  
\begin{document}
\maketitle
\begin{abstract}
Recent advancements in large language models have revolutionized text generation with their remarkable capabilities. These models can produce controlled texts that closely adhere to specific requirements when prompted appropriately. However, designing an optimal prompt to control multiple attributes simultaneously can be challenging. A common approach is to linearly combine single-attribute models, but this strategy often overlooks attribute overlaps and can lead to conflicts. Therefore, we propose a novel combination strategy inspired by \emph{the Law of Total Probability} and \emph{Conditional Mutual Information Minimization} on generative language models. This method has been adapted for single-attribute control scenario and is termed the \textbf{Palette of Language Models} due to its theoretical linkage between attribute strength and generation style, akin to blending colors on an artist's palette. Moreover, \emph{positive correlation} and \emph{attribute enhancement} are advanced as theoretical properties to guide a rational combination strategy design. We conduct experiments on both single control and multiple control settings, and achieve surpassing results.
\end{abstract}

\section{Introduction}

The purpose of controlled text generation is to modify the output of the language models with a pre-given attribute, so that the final output conforms to the attribute \cite{pmlr-v70-hu17e,madaan2021generate,zhang2023survey}. It is common to utilize Bayes Rules \cite{li2022diffusionlm} to modify the language model $p(X)$ to form with the conditional variable $p(X|a)$, and by virtue of the corresponding discriminative model (generally a classification model), implement the constraints on the generated results. This approach will face two problems: On the one hand, the discriminative model usually needs to be fine-tuned according to the generation task scenario to better assist controlled text generation, because the accuracy of the discriminative model plays a key role in the generation effect. Nevertheless, it is time-consuming to collect the corresponding classification data. On the other hand, the classification effect of the discriminative model often depends on certain words or phrases related to attributes, therefore, in the process of predicting the next token, the discriminative model will guide the language model to bias these words, which makes the generated results lack diversity.

In recent years, with the rapid development of large language models \cite{chowdhery2022palm,du2022glam,hoffmann2022training}, models represented by ChatGPT \cite{Achiam2023GPT4TR} have powerful text generation capabilities. Many generation tasks can be converted to prompt engineering to use large language models to get better solutions. Similarly, attributes can be designed as prompts, so that with the powerful generation ability of the large language model, the final generated text will also conform to such attributes. When the number of attributes that need to be met is large, it is difficult to design a suitable prompt to cover these attributes. Also, due to the inherent ambiguity and sensitivity of prompts, we can't guarantee that the generated results will be perfectly reproduced according to the attributes mentioned in prompts.

Model Arithmetic \cite{dekoninck2024controlled} proposes a framework to ensemble multiple attributes, in which through simple arithmetic operations, such as linear composition, multiple attribute-related language models or discriminative models are combined to obtain a multi-attribute controlled text generation strategy. This framework treats the language models associated with different attributes as independent. However, in reality, each language model may have multiple attributes, and the attributes between language models may overlap which can not be effectively modeled with simple arithmetic operations. For example, the main attribute of language model A is "formal response", and that of language model B is "child's tone" (which suggests an attribute of "informal response" as well). With linear combination, the attribute of "formal response" is faded, making the final generation strategy incomplete.

Deriving from \emph{the Law of Total Probability} and \emph{Conditional Mutual Information Minimization}, we propose \textbf{Palette of Language Models}, which alleviates the latent attribute overlapping problem when attributes assembled. The improved combination strategy has following contributions:

$\bullet$ Leveraging \emph{the Law of Total Probability}, we decompose the final generation distribution as attribute satisfied event and its complementary event of which the latter never appears in previous works.

$\bullet$ We model the attributes overlapping problem as \emph{Conditional Mutual Information Minimization}, with which a dynamic coefficient on each attribute (as a part of attribute strength) is derived for attribute enhancement. We also take theoretical analysis on attribute strength and conclude the positive correlation between it and the final generation style.

$\bullet$ We propose two pieces of theoretical properties (\emph{positive correlation} and \emph{attribute enhancement}) which guide a rational attribute combination.  

\section{Related work}

Some works for controlling language models are training or aligning language models conditioned on static or iterative updated control codes that make desired features of generated text more explicit \cite{keskar2019, DBLP:journals/corr/abs-2005-00558, NEURIPS2022_b125999b}. These methods often depend on a large amount of training data, which will result in a lot of annotation costs and resource consumption. 
In order to solve the above problems, some researchers have proposed methods of fine-tuning \cite{2021On,yuan2023large}, prompt tuning \cite{li2022learning,pagnoni2021understanding} or Prefix Tuning \cite{qian-etal-2022-controllable,clive2022control,ma-etal-2023-focused} large models to generate controlled text. However, due to the use of GPT series models, it will cause infeasible and cannot solve the toxicity and bias problems of the model \cite{tonmoy2024comprehensive}.
Although Timo Schick et al. \cite{schick2021selfdiagnosis} found that pre-trained models can largely recognize their bad biases and the toxicity of the content they produce, by giving textual descriptions of bad behavior when decoding, the probability of generating problematic text can be reduced. But this method is greedy in nature when supporting or opposing a decision must always use a specific word, taking into account only the context in which it has been generated.

Another category of method uses gradients of single or combination of attribute discriminators based on energy to update generating language model’s hidden states to guide decoding process \cite{DBLP:conf/iclr/DathathriMLHFMY20, DBLP:conf/acl/MireshghallahGB22, DBLP:conf/nips/QinWKC22, DBLP:conf/emnlp/KumarPT22, DBLP:conf/nips/KumarMST21}, which do not need labeled training data. Despite the use of Langevin dynamics, samples the sequence of token embeddings instead of logits and Gibbs sampling methods. The main issue of this category is that multiple sampling iterations are required to converge, which slow down the decoding progress. BOLT \cite{liu2023bolt}, instead, preserves token dependencies and simultaneously optimizes via autoregressive decoding to constrain by adding biases. But BOLT requires stricter constraints, for example, keyword control requires more than three keywords. BOLT also requires careful tuning of the different hyperparameters constituting the energy function—a problem prevalent in energy-based controlled power generation approaches.

Some methods are inspired by Bayes Rules that use attribute probability from discriminators to steer language generation towards desired attributes. The attribute discriminators can be discriminative or generative. FUDGE \cite{DBLP:conf/naacl/YangK21} learns a binary discriminator for whether attribute will become true in future, and the output probabilities of this discriminator are multiplied with generator’s original probabilities to get the desired probabilities. Gedi \cite{krause-etal-2021-gedi-generative} uses class-conditional language models as generative discriminators, which results in faster computation speed due to the parallel computation of all candidate tokens. Compared with these methods, PREADD \cite{DBLP:conf/acl/PeiYK23} considers the language generator with a prefix-prepended prompt as the role of attribute discriminator, which does not require an external model and corresponding additional training data.

Recent work proposes controlling text generation via language model arithmetic \cite{dekoninck2024controlled}, which enables to combine multiple language models of different attributes into a formula-based composite. The key distinguishing feature of our method is that we explicitly model the attributes overlapping between language models, so that they are not faded during combination. 

\begin{figure*}
    \centering
    \includegraphics[width=1.0\textwidth]{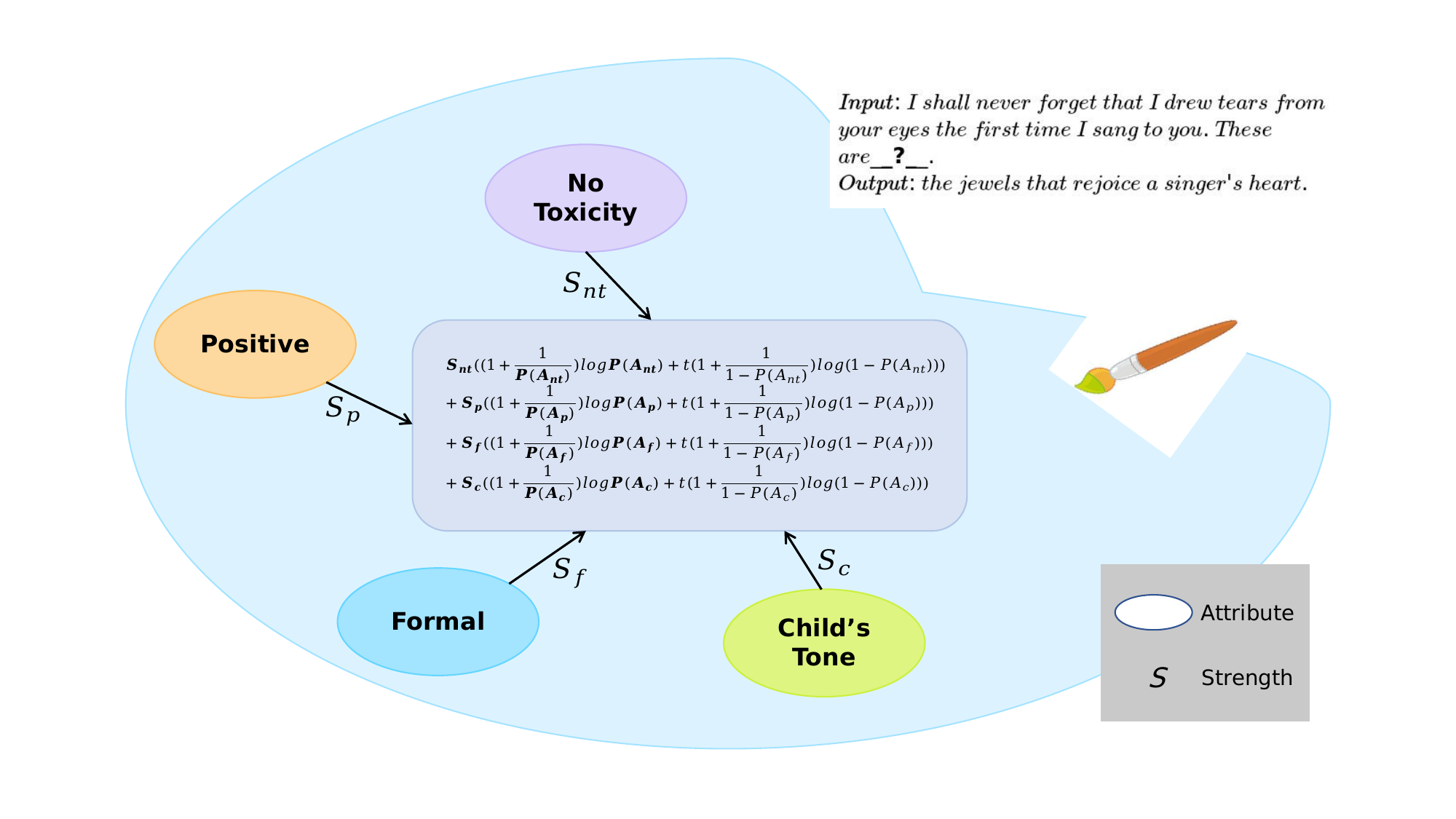}
    \caption{Overview of \textbf{Palette of Language Models}. Each ellipse in the figure represents a generative language model with a specific attribute, and $S$ represents the strength of the corresponding model. Employing Equation \ref{equ:comb_norm}, the final generation under multiple constraints is derived.}
    \label{fig:palettel}
\end{figure*}

\section{Proposed method}
\label{sec:method}
In this paper, we focus on \textbf{n} (where $n >= 2$) attributes combination strategy to derive a specific output distribution on controlled text generation which indicates a mixture style. Analogously, single attribute control can be treated as the combination with the basic model. We attach each attribute to a generative language model for ease of the formula derivation.

\subsection{Problem definition}
\label{sec:problem}
Assuming \textbf{n} generative language models, with each owning a certain attribute, such as "able to generate text with positive sentiment", "speak in a child's tone", or "tell a topic about the solar system", etc. For the \textbf{i}th model $A_{i}$, the prediction probability of the next token is $p_{A_{i}}(x_{t}=x|x_{1:t-1})$, abbreviated as $p(A_{i}=x)$ (where $x \in V$, the vocabulary of current language model). Thus, we desire a function $\mathcal{F}$ on $p(A_{i})$ and express the final distribution as:
\begin{gather}
    p(Z) = \mathcal{F}(p(A_{1}),...,p(A_{i}),...,p(A_{n})) \notag\\
    \textbf{s.t.} \hspace{4pt} 
    p(A_{i},A_{j}) = p(A_{i})p(A_{j}), \notag \\ min(\mathcal{M}(A_{i}, A_{j})) \hspace{4pt} (for \hspace{2pt} 1 \leq i,j \leq n)
    \label{equ:problem}
\end{gather}
where $\mathcal{M}(.)$ is a metric to gauge the overlapping between an attribute couple under the final distribution. It is noteworthy that the joint distribution item, i.e., $p(A_{i}, A_{j})$, means the probability that “the next predicated tokens for both attributes $A_{i}$ and $A_{j}$ are x”. Obviously, they are separate neural networks and none inter-influence exists, which ensures the independence between them.

For the attribute couple $(A_{i},A_{j})$, the \emph{conditional mutual information} under the final distribution $Z$ is always no less than 0. And \textbf{if it is greater than 0, the overlapping between them happens}. Therefore, the overlapping metric $\mathcal{M}(.)$ mentioned in Equation \ref{equ:problem} is defined as:
\begin{gather}
    \mathcal{M}(A_{i},A_{j})=\textbf{I}(A_{i},A_{j}|Z) = \notag \\
    \sum_{z}p(z)\!\!\sum_{a_{i}, a_{j}} p(a_{i},a_{j}|z)log\frac{p(a_{i},a_{j}|z)}{p(a_{i}|z)p(a_{j}|z)} \geq 0
    \label{equ:overlapping}
\end{gather}
where $z \in Z$, $a_{i}\in A_{i}$ and $a_{j}\in A_{j}$ are satisfied.

\subsection{Proposed solution}
\label{sec:solution}
 For the final generation distribution, i.e., $p(Z=x)$, we use \emph{the Law of Total Probability}, which, in the case of a single attribute $A_{i}$, can be written as:
\begin{gather}
    p(Z=x) = \lambda_{i}*p(A_{i}=x) + 
   \lambda_{i^{'}}*p(A_{i} \neq x) \notag \\
   \textbf{s.t.} \hspace{4pt}\lambda_{i} = p(Z=x|A_{i}=x), \notag\\
\lambda_{i^{'}} =  p(Z=x|A_{i} \neq x)
    \label{equ:single_factorization}
\end{gather}
Similarly, for the attribute couple $(A_{i}, A_{j})$, taking into account the independence between them (see Equation \ref{equ:problem}), the final generation strategy can be written as:
\begin{gather}
    p(Z=x)=\lambda_{ij}*p(A_{i}=x)p(A_{j}=x) \notag \\
    +\lambda_{i^{'}j^{'}}*p(A_{i} \neq x)p(A_{j}\neq x)\notag\\
    +\lambda_{ij^{'}}*p(A_{i}=x)p(A_{j}\neq x) \notag \\
    +\lambda_{i^{'}j}*p(A_{i}\neq x)p(A_{j}=x) \notag \\
    \textbf{s.t.} \hspace{4pt} \lambda_{ij}=p(Z=x|A_{i}=x,A_{j}=x), \notag \\
    \lambda_{i^{'}j^{'}}=p(Z=x|A_{i}\neq x,A_{j} \neq x),\notag \\
    \lambda_{ij^{'}}=p(Z=x|A_{i}=x,A_{j} \neq x), \notag \\
    \lambda_{i^{'}j}=p(Z=x|A_{i}\neq x,A_{j}=x)
    \label{equ:dual_factorization}
\end{gather}

Employing the convexity of the logarithmic function, i.e., $log(ax+by) - log(a+b)\geq \frac{a}{a+b}log(x)+\frac{b}{a+b}log(y)$, we add both single and couple factorization equations aforementioned, and obtain an approximate expression (details are shown in Section \ref{sec:ac_detail}):
\begin{gather}
    log3p(Z) \approx \sum_{s \in \{i,j\}}\alpha_{s}logp(A_{s} = x) \notag \\
    + \sum_{s \in \{i,j\}}\beta_{s}logp(A_{s}\neq x), \notag \\
    \textbf{s.t.} \hspace{4pt} \alpha_{i} \!\!=\!\! \lambda_{i}\!\!+\!\!\lambda_{ij}\!\!+\!\!\lambda_{ij^{'}}, \hspace{4pt}
    \beta_{i} \!\!=\!\! \lambda_{i^{'}}\!\!+\!\!\lambda_{i^{'}j}\!\!+\!\!\lambda_{i^{'}j^{'}}
    \label{equ:approx_couple}
\end{gather}
Additionally, by minimizing the conditional mutual information in Equation \ref{equ:overlapping}, we 
derive that (details are shown in Section \ref{sec:pr_ci}):
\begin{gather}
\lambda_{ij}=\frac{\lambda_{i}\lambda_{j}}{p(Z=x)},\lambda_{i^{'}j^{'}}=\frac{\lambda_{i^{'}}\lambda_{j^{'}}}{p(Z=x)},\notag\\
\lambda_{ij^{'}}=\frac{\lambda_{i}\lambda_{j^{'}}}{p(Z=x)},\lambda_{i^{'}j}=\frac{\lambda_{i^{'}}\lambda_{j}}{p(Z=x)}
\label{equ:couple_rel}
\end{gather}

We traverse the pairwise combinations of $\textbf{n}$ attributes (altogether \textbf{$C_{n}^{2}$} items) and add them in the form of Equation \ref{equ:approx_couple}:
\begin{gather}
logp(Z=x) \propto \sum_{i=1}^{n} \varphi_{i}logp(A_{i} = x) \notag \\ 
+ \sum_{i=1}^{n} 
 \omega_{i}logp(A_{i} \neq x) \notag \\ 
\textbf{s.t.} \hspace{4pt} \varphi_{i} \!\!=\!\! \frac{(n\!\!-\!\!1)\lambda_{i}\!\!+\!\!\frac{p(A_{i}=x|Z=x)}{p(A_{i}=x)}\sum_{j \neq i}(\lambda_{j}\!\!+\!\!\lambda_{j^{'}})}{C_{n}^{2}}, \notag \\
\omega_{i} \!\!=\!\! \frac{(n\!\!-\!\!1)\lambda_{i^{'}}\!\!+\!\!\frac{p(A_{i}\neq x|Z=x)}{p(A_{i} \neq x)}\sum_{j \neq i}(\lambda_{j}\!\!+\!\!\lambda_{j^{'}})}{C_{n}^{2}}
\label{equ:improve_linear}
\end{gather}

Inspired by previous works, we introduce the generative language model with standard output (which means no bias to any specific attributes) as the basic part in aid of generation stability, and simplify the expression in Equation \ref{equ:improve_linear} (details are shown in Section \ref{sec:simp}). Therefore, both $single$ and $multiple$ attributes can achieve the final generation solution in a consolidated manner (where the $single$ setting can be treated as the combination with the basic part):
\begin{gather}
    logp(Z = x) \propto \frac{\sum\limits_{i=1}^{n} s_{i}c_{i}logp(A_{i} = x) + logP_{b}}{M_{1}} \notag \\
    + t \hspace{4pt} \frac{\sum\limits_{i=1}^{n} s_{i^{'}}c_{i^{'}}logp(A_{i} \neq x) }{M_{2}} \notag \\
    \textbf{s.t.} \hspace{4pt} c_{i}=1\!\!+\!\!\frac{1}{p(A_{i}=x)}, \hspace{4pt} c_{i^{'}}=1\!\!+\!\!\frac{1}{p(A_{i} \neq x)}
    \label{equ:comb_norm} 
\end{gather}
where $M_{1}$ and $M_{2}$ are normalization values with which the combination logits will be as the same order of magnitude as the basic language model. $P_{b}$ is the generation probability of the basic part. $s_{i}$ and $s_{i^{'}}$ are variables to express the strength (the proportion to the final generation) of the current attribute. $t$ is a small coefficient which servers the complementary event, i.e., $p(A_{i} \neq x)$, as an auxiliary for the final generation strategy.

\subsection{Strategy properties}
\label{sec:propery}
\paragraph{Property 1.}\emph{It demonstrates a positive correlation between attribute strength, i.e., $s_{i}$, and the final generation style about our improved attribute combination strategy.} (The proof is shown in Section \ref{sec:psp})

Specifically, we introduce the attribute token $x_{attr} \in V$ (vocabulary of the language model) which can better express the property of current model, hence:
\begin{equation}
    p(A_{i}=x_{attr}) > p(A_{i}=x_{t}) \hspace{4pt} (\forall t\neq attr)
    \label{equ:attr_token}
\end{equation}
\paragraph{Property 2.}\emph{Our strategy gets attribute enhancement compared to linear combination strategy.} (The proof is shown in Section \ref{sec:psp})

As for two combination strategies $\mathcal{F}_{1}$ and $\mathcal{F}_{2}$, the attribute $A_{i}$ get enhancement from $\mathcal{F}_{1}$ to $\mathcal{F}_{2}$, if:
\begin{equation}
    p(Z_{2}=x_{attr}) > p(Z_{1}=x_{attr})
    \label{equ:att_enh}
\end{equation}
where $p(Z_{k})$ means the final generation probability over strategy $\mathcal{F}_{k}$, $x_{attr}$ is the attribute token of $A_{i}$.
\section{Experiments}
\label{sec:experiment}

\begin{table*}[ht!]
  \centering
    \begin{tabular}{cc}
    \toprule  
    \textbf{Attribute} & \textbf{Prompt}\\
    \midrule  
    \textbf{reducing toxicity} & \makecell[c]{The following conversation is one that does not perpetuate \\ negative stereotypes, is not threatening and does not contain \\any sexually explicit or profane language.}\\
    \hline
    \textbf{enabling positive sentiment} &\makecell[c]{The following conversation is one that is happy and positive.}\\
    \hline
    \textbf{enabling fluency} &\makecell[c]{Continue writing after the conversation to make it more fluency \\ and readable.}\\
    \bottomrule 
    \end{tabular}%
    \caption{Attributes and prompts for toxicity reduction.}
  \label{tab:toxi_prompt}%
\end{table*}%

\begin{table*}
\centering  
\begin{tabular}{ccccccc}
\toprule                  
{} & \multicolumn{2}{c}{Llama2-7b} &\multicolumn{2}{c}{Pythia-12b}&\multicolumn{2}{c} {MPT-7b}\\
{} & {Tox.} & {Perpl.}& {Tox.}&{Perpl.}& {Tox.} &{Perpl.}\\
\hline
{\textbf{M}}&{0.270} &{12.87} &{0.250} &{21.29} &{0.266} &{\textbf{18.61}}\\ 
\hline
{\textbf{Self-Debiasing }($\lambda=10$)}&{0.257} &{14.25} &{0.241} &{25.37} &{0.268} &{21.19}\\ 
{\textbf{FUDGE }($M+C$)}&{0.233} &{13.56} &{0.208} &{21.80} &{0.225} &{19.53}\\ 
{\textbf{PREADD }($M\!\!-\!\!0.6M_{toxic}$)}&{0.215} &{12.66} &{\textbf{0.176}} &{32.88} &{\textbf{0.200}} &{23.01}\\
{\textbf{Linear }($M\!\!-\!\!0.96 union(M_{toxic},M)$)}&{0.199} &{10.61} &{0.179} &{22.74} &{0.204} &{19.32}\\ 
\hline
{\textbf{Ours}}&{\textbf{0.159}}&{\textbf{9.42}}&{0.186}&{\textbf{20.34}}&{0.213}&{24.07}\\
\bottomrule
\end{tabular}
\caption{Toxicity and perplexity of various methods on the {/pol/} dataset. $M$ and $M_{toxic}$ denotes the methods without/with conditioning to toxicity respectively. $C$ is a toxicity classifier. Perplexity is measured with respect to \textbf{M}. Lower is better.}
\label{table:toxi_eval}
\end{table*}

We evaluate the improved attribute combination strategy in both $single$ and $multiple$ control scenarios to testify the attribute enhancement. Specifically, Sections \ref{sec:tox_red} and \ref{sec:sent_control} are for the single setting, and the Section \ref{sec:multi_control} is for multi-attribute control which also demonstrates the overlapping alleviation. We conduct experiment for positive correlation and complementary event verification in Sections \ref{sec:corr}. All the experiments are implemented on a single GPU of \emph{Tesla V100S-PCIE-32GB}.

\paragraph{Base models.} We conduct experiments on several popular autoregressive large language models for strategy evaluation: Llama2-7b\footnote{\url{https://huggingface.co/meta-llama/Llama-2-7b}}\cite{Touvron2023Llama2O}, Pythia-12b\footnote{\url{https://huggingface.co/EleutherAI/pythia-12b}}\cite{Biderman2023PythiaAS} and MPT-7b\footnote{\url{https://huggingface.co/mosaicml/mpt-7b}}\cite{MosaicML2023Introducing}.

\paragraph{Baselines for comparison.} The baselines we make comparison with are as follows: \textbf{M}(the basic language model without any prompt), \textbf{Self-Debiasing \cite{schick2021selfdiagnosis}}, \textbf{FUDGE \cite{DBLP:conf/naacl/YangK21}}, \textbf{PREADD \cite{DBLP:conf/acl/PeiYK23}} and \textbf{Linear combination \cite{dekoninck2024controlled}} which is derived from the KL-Optimality and satisfies:
\begin{equation}
    log p(Z=x) \propto \sum_{i=1}^{n} \lambda_{i} log p(A_{i}=x)
    \label{equ:linear_comb}
\end{equation}
Linear combination is a concise strategy to merge multiple attributes, and makes operation on $\lambda_{i}$ to bias the overall output toward ($\lambda_{i}>0$) or away from ($\lambda_{i}<0$) the attribute $A_{i}$. However, it does neglect the overlapping between attributes which might cause attribute conflict when combination.

As for the linear combination method, we employ its best ARITHMETIC strategy, $M\!\!-\!\!0.96Union(.)$, for comparison.

\begin{table*}
\centering  
\resizebox{\linewidth}{!}{
\begin{tabular}{ccccccc}
\toprule 
\multirow{2}{*}{\textbf{Negative $\longrightarrow$ Positive}} &\multicolumn{2}{c}{Llama2-7b} &\multicolumn{2}{c}{Pythia-12b}&\multicolumn{2}{c} {MPT-7b}\\
{} & {Sentiment.} & {Perpl.}& {Sentiment.}&{Perpl.}& {Sentiment.} &{Perpl.}\\
\hline
{\textbf{M}}&{0.218} &{14.15} &{0.212} &{22.86} &{0.210} &{19.41}\\ 
{$M_{pos}$}&{0.239} &{13.69} &{0.244} &{21.69} &{0.244} &{18.25}\\ 
\hline
{\textbf{Self-Debiasing }($\lambda=10$)}&{0.270} &{14.99} &{0.244} &{25.17} &{0.251} &{19.79}\\ 
{\textbf{FUDGE}($M+C$)}&{0.339} &{13.73} &{0.337} &{22.63} &{0.326} &{18.81}\\ 
{\textbf{PREADD }($M\!\!-\!\!0.6M_{pos}$)}&{0.373} &{14.20} &{\textbf{0.343}} &{30.86} &{0.343} &{19.25}\\
{\textbf{Linear }($M_{pos}\!\!-\!\!0.96 union(M_{pos},M_{neg})$)}&{0.411} &{\textbf{12.82}} &{0.343} &{22.53} &{0.370} &{\textbf{17.88}}\\ 
\hline
{\textbf{Ours}}&{\textbf{0.426}}&{19.42}&{0.336}&{\textbf{15.33}}&{\textbf{0.405}}&{18.85}\\
\Xhline{1.5pt}               
\multirow{2}{*}{\textbf{Positive $\longrightarrow$  Negative}} &\multicolumn{2}{c}{Llama2-7b} &\multicolumn{2}{c}{Pythia-12b}&\multicolumn{2}{c} {MPT-7b}\\
{} & {Sentiment.} & {Perpl.}& {Sentiment.}&{Perpl.}& {Sentiment.} &{Perpl.}\\
\hline
{\textbf{M}}&{0.204} &{13.51} &{0.218} &{\textbf{22.25}} &{0.196} &{18.26}\\ 
{$M_{neg}$}&{0.299} &{14.17} &{0.340} &{22.73} &{0.284} &{17.83}\\ 
\hline
{\textbf{Self-Debiasing }($\lambda=10$)}&{0.339} &{15.02} &{0.364} &{25.59} &{0.285} &{20.02}\\ 
{\textbf{FUDGE }($M+C$)}&{0.410} &{14.97} &{0.433} &{23.39} &{0.377} &{18.43}\\ 
{\textbf{PREADD }($M\!\!-\!\!0.6M_{neg}$)}&{0.470} &{14.66} &{\textbf{0.514}} &{32.03} &{0.438} &{19.44}\\
{\textbf{Linear }($M_{neg}\!\!-\!\!0.96 union(M_{neg},M_{pos})$)}&{0.502} &{\textbf{13.12}} &{0.499} &{24.93} &{0.452} &{18.23}\\ 
\hline
{\textbf{Ours}}&{\textbf{0.547}}&{16.31}&{0.469}&{30.41}&{\textbf{0.478}}&{\textbf{15.81}}\\
\bottomrule
\end{tabular}
}
\caption{Sentiment score and Perplexity value of various methods on the {IMDB} dataset for \textbf{"negative to positive"} and \textbf{"positive to negative"} transition tasks.
$M_{pos}$ and $M_{neg}$ denote the models with conditioning to positive/negative sentiments respectively. $C$ is a sentiment classifier. Perplexity is measured with respect to \textbf{M}. Lower is better.}
\label{table:senti_transition}
\end{table*}

\paragraph{Experiment Setting.} We notice that implementation of the standard normalization $M$ in Equation \ref{equ:comb_norm} remains elusive due to the fact that the variable $c_{i}$ tends to infinite if $p(A_{i}=x)$ is trivial. With superseding it, we introduce $sigmoid$ function ($\sigma(p(A_{i})=x)$) for its keeping the major properties in Section \ref{sec:propery} (details are shown in Section \ref{sec:sigma_proof}). In addition, instead of training language models with a specific attribute from scratch, we induce the attribute in the basic language model with the \textbf{prompt} engineering. Hence, variables for experiments of the improved strategy are:
\begin{gather}
    c_{i} = 1 \!\!+\!\! \frac{1}{\sigma(p(A_{i} = x)}, \hspace{4pt} c_{i^{'}} = 1 \!\!+\!\! \frac{1}{\sigma(p(A_{i} \neq x)},\notag \\
    M_{1} \!\!=\!\! 1 \!\!+\!\! (2+\frac{1}{e})\sum_{i=1}^{n}s_{i}, \hspace{4pt} M_{2} \!\!=\!\! (2+\frac{1}{e})\sum_{i=1}^{n}s_{i^{'}}, \notag \\
    \hspace{4pt} p(A_{i}=x)=P_{b}(x|prompt_{A_{i}})
    \label{equ:comb_norm_exper}
\end{gather}
where $prompt_{A_{i}}$ are displayed in Tables \ref{tab:toxi_prompt}, \ref{tab:senti_prompt}, \ref{tab:prompts_overlapping} and \ref{tab:prompts_overlapping_same}.

\subsection{Toxicity reduction}
\label{sec:tox_red}
We first test our algorithm in terms of toxicity reduction on \textbf{/pol/} dataset \cite{osti_10212018}, which comprises contents from website 4chan\footnote{\url{https://boards.4chan.org/pol/}} and attaches each item a toxicity score. We randomly select 2000 samples with their scores greater than 0.5. For toxicity reduction process, we construct a dialogue stuff pattern in which original toxicity texts are from the inquirer (i.e., \textbf{Person 1:}), and with that, the attribute mixture with combined strategy (as \textbf{Person 2:}) is compelled to generate toxicity-free utterance. We assemble three attributes of which the positive sentiment and fluency are as supplementaries for toxicity reduction. Each attribute and its corresponding prompt are in Table \ref{tab:toxi_prompt}. The metrics picked in this setting are \textbf{Toxicity Score} (which measures the virulent degree by Perspective API\footnote{\url{https://perspectiveapi.com/}}) and \textbf{PPL} (which estimates generation consistency according to perplexity calculation).

As is shown in Table \ref{table:toxi_eval}, with Llama2-7b as the basic language model, both toxicity score and PPL value are at the first-rate where the toxicity probability degrades to a high-quality level with $4\%$ compared to traditional SOTA methods, which embodies advancement of attribute enhancement in our algorithm. The proposed strategy perform a slight inferior (about $1\%$ at a disadvantage in toxicity) than its comparators under Pythia-12b and MPT-7b settings.

\subsection{Sentiment control}
\label{sec:sent_control}
\begin{table}[ht!]
  \centering
    \begin{tabular}{c|c}
    \toprule
    \textbf{Attribute} & \textbf{Prompt}\\
    \hline
    \makecell[c]{\textbf{positive}\\ \textbf{reply}} & \makecell[c]{The following is a positive movie\\review, with a very positive\\sentiment and a very positive tone.}\\
    \hline
    \makecell[c]{\textbf{negative}\\ \textbf{reply}} & \makecell[c]{The following is a negative movie\\review, with a very negative\\sentiment and a very negative tone.}\\
    \bottomrule
    \end{tabular}%
    \caption{Attributes and prompts for sentiment transition.}
  \label{tab:senti_prompt}%
\end{table}%

\begin{figure*}[h]
\centering
\includegraphics[width=1.0\textwidth]{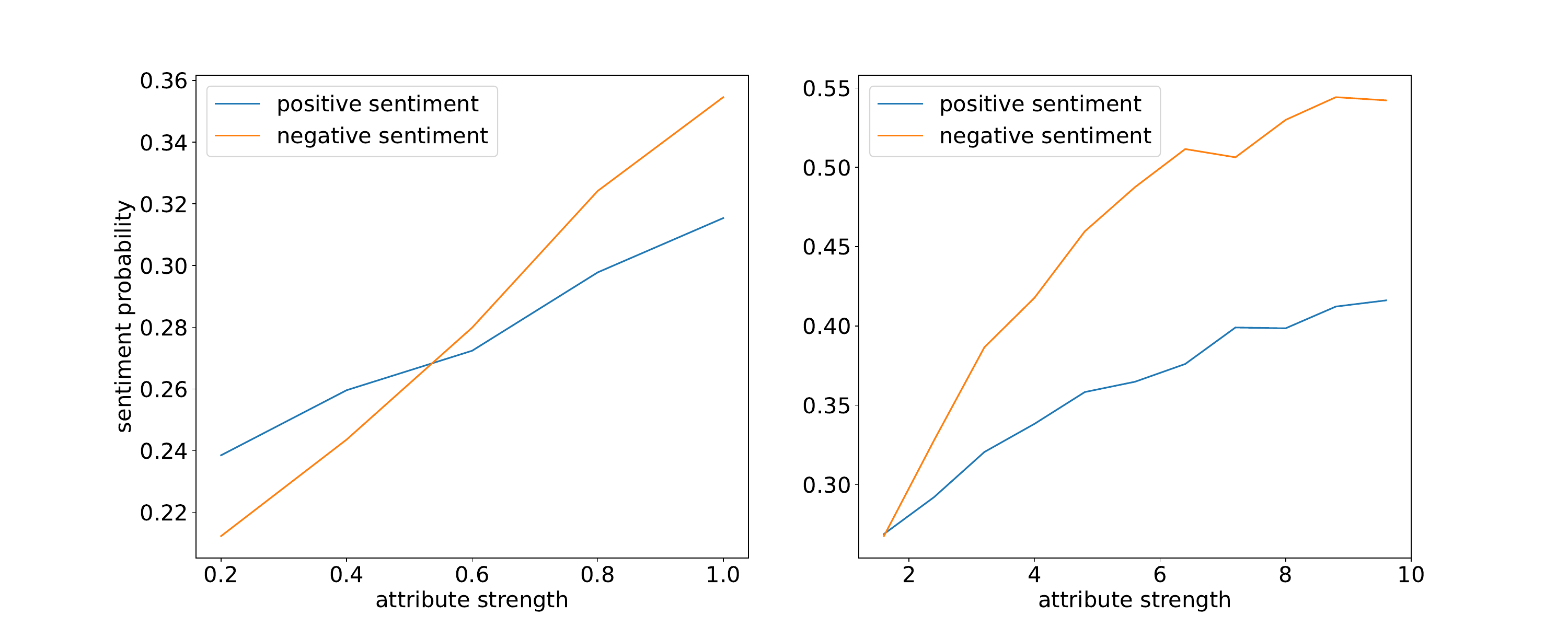}
\caption{Positive Correlation between attribute strength \& sentiment score (Left: $s\!\!<\!\!1$, Right: $s\!\!>\!\!1$).}
\label{fig:pos_corr}
\end{figure*}

We make a subset of \textbf{IMDB movie review} dataset \cite{maas-etal-2011-learning} with 1000 samples separately for both positive and negative sentiment control. Following the setting of \citeauthor{dekoninck2024controlled}, we also keep the first 32 tokens of original sentences retained and force language models to write after them with an opposite sentiment. This setting demonstrates hostile performance for language models owing to their generally decoding in complying with previous tokens (in both structural and semantic consistency). Inspired by \cite{liu-etal-2021-dexperts,dekoninck2024controlled}, we enhance the accent on required sentiment via deducting logits of its antagonistic stuff. That is:
\begin{gather}
        logp(Z=x) \!\! \propto \!\!\frac{\sum f(i)\!\!*\!\!s_{i}c_{i}logp(A_{i} \!\! = \!\! x) + logP_{b}}{M_{1}} \notag\\ 
        + t \frac{\sum f(i)\!\!*\!\!s_{i^{'}}c_{i^{'}}logp(A_{i} \!\! \neq \!\! x)}{M_{2}}\notag \\
        \textbf{s.t.} \hspace{4pt} i \in \{main, anti\}, f(i)=
        \left\{
        \begin{aligned}
        1 \quad main \\
        -1 \quad anti
        \end{aligned}
        \right
        .
    \label{equ:sent_eval}
\end{gather}

For instance, if the required sentiment is \textbf{positive}, then \textbf{main} is for positive attribute and \textbf{anti} is for negative. The prompts for both positive and negative sentiments are enumerated in Table \ref{tab:senti_prompt}. Following \citeauthor{dekoninck2024controlled}, we take the \textbf{twitter-sentiment} discriminative model\footnote{\url{https://huggingface.co/cardiffnlp/twitter-roberta-base-sentiment-latest}} for sentiment score derivation.

Referring to Table \ref{table:senti_transition}, with both Llama2-7b and MPT-7b as the basic models, our strategy exceeds other methods on the metric of sentiment score. Specifically, in Llama2-7b setting, we obtain an average score of $3\%$ ($1.5\%$ for neg2pos and $4.5\%$ for pos2neg transition) greater than the SOTA method. The values are $3.5\%$ (neg2pos) and $2.6\%$ (pos2neg) in MPT-7b setting. As for PPL value, our strategy demonstrates a modicum of weakness at a manageable level in which the average relative values on the SOTA method are $+4.9$, $+0.9$ and $-0.5$ for the three basic models. Therefore, without sacrificing the generation quality, our algorithm gains attribute enhancement in the sentiment control setting.


\subsection{Evaluations on attribute strength $s_{i}$ and the complementary event $A_{i} \neq x$}
\label{sec:corr}
We verify the positive correlation property between attribute strength $s$ and the generation style (see \textbf{Property 1} in Section \ref{sec:propery}) in this part. Following the setting of Section \ref{sec:sent_control}, we dynamically vary strength of the main attribute $s_{main}$ in Equation \ref{equ:sent_eval} and chronicle sentiment scores. The basic model we choose is Llama2-7b. From Figure \ref{fig:pos_corr}, we judge that the final generation performance inclines to ascend on account of gaining strength in both cases that $s<1$ and $s>1$.

As is processed in Equation \ref{equ:comb_norm}, the complementary event $A_{i} \neq x$ is treated as the auxiliary for the main part $A_{i}=x$. We make evaluations of the impact it makes on the final generation style. Similar to Section \ref{sec:corr}, we also conduct experiments on sentiment control with Llama2-7b and vary the coefficient $t$ of $logp(A_{i} \neq x)$ in Equation \ref{equ:sent_eval}.

Shown in Figures \ref{fig:pos_t} and \ref{fig:neg_t} (see in Section \ref{sec:ce_pic}), the dash lines represent sentiment probabilities with $t=0$ (which equal to $0.4161$ and $0.5442$ in positive \& negative settings, separately). From both figures, we discover that the final generation performance fluctuates nearby the dash line with the value of $t$.

\begin{table*}[ht!]
  \centering
    \begin{tabular}{ccc}
    \toprule
    \textbf{Settings} & \textbf{Attribute} & \textbf{Prompt}\\
    \hline
    \multirow{2}{*}{\textbf{\makecell[c]{Setting \\one}}} & 
    \textbf{toxicity} & \makecell[c]{The following movie review is not threatening and does not contain any\\sexually explicit or profane language, and is with a slight \textbf{negative} sentiment.} \\ 
    \cline{2-3}
    & \textbf{\makecell[c]{positive \\reply}} & \makecell[c]{Continue writing after the movie review to change it with \\ a very \textbf{positive} sentiment and a very \textbf{positive} tone.}\\
    \hline
    \multirow{2}{*}{{\textbf{\makecell[c]{Setting \\two}}}} & 
    \textbf{toxicity} & \makecell[c]{The following movie review is not threatening and does not contain any\\sexually explicit or profane language, and is with a slight \textbf{positive} sentiment.} \\ 
    \cline{2-3}
    & \textbf{\makecell[c]{negative \\reply}} & \makecell[c]{Continue writing after the movie review to change it with \\a very \textbf{negative} sentiment and a very \textbf{negative} tone.}\\
    \bottomrule
    \end{tabular}%
    \caption{Attributes and prompts for the multi-control with overlapping setting.}
  \label{tab:prompts_overlapping}%
\end{table*}%

\begin{table*}[h!]
  \centering
  \resizebox{\linewidth}{!}{
    \begin{tabular}{ccccccccccccc}
    \toprule
    \multicolumn{2}{c}{\textbf{Positive }} & \textbf{0.1} & \textbf{0.2} & \textbf{0.3} & \textbf{0.4} & \textbf{0.5} & \textbf{0.6} & \textbf{0.7} & \textbf{0.8} & \textbf{0.9} & \textbf{1.0} & \textbf{Average} \\
    \hline
    \multirow{3}{*}{\makecell*[c]{\textbf{Perplexity($\downarrow$)}}} & \textbf{Linear} &{14.95} &{15.48} &{15.36} &{15.84} &{15.70} &{15.17} &{15.71} &{15.57} &{15.51} &{15.86} &{15.51}  \\ 
    \cline{2-13}
     & \textbf{Union}
    &{13.72} &{14.25} &{13.50} &{13.97} &{13.82} &{14.16} &{13.71} &{13.98} &{13.95} &{13.90} &{13.90} \\
    \cline{2-13}
    &\textbf{Ours}
    &\textbf{6.40} &\textbf{6.54} &\textbf{6.63} &\textbf{6.52} &\textbf{6.60} &\textbf{6.55} &\textbf{6.66} &\textbf{6.42} &\textbf{6.60} &\textbf{6.68} &\textbf{6.56} \\
    \hline
    \multirow{3}{*}{\makecell*[c]{\textbf{Sentiment($\uparrow$)}}} & \textbf{Linear}
    &{0.543} &{0.530} &{0.539} &{0.533} &{0.526} &{0.535} &{0.534} &{0.525} &{0.531} &{0.534} &{0.533} \\
    \cline{2-13}
    & \textbf{Union}
    &{0.443} &{0.439} &{0.441} &{0.445} &{0.437} &{0.444} &{0.439} &{0.443} &{0.446} &{0.440} &{0.442} \\
    \cline{2-13}
    &\textbf{Ours}
    &\textbf{0.565} &\textbf{0.562} &\textbf{0.556} &\textbf{0.543} &\textbf{0.559} &\textbf{0.564} &\textbf{0.558} &\textbf{0.551} &\textbf{0.557} &\textbf{0.544} &\textbf{0.556}  \\
    \Xhline{1.5pt}
    \multicolumn{2}{c}{\textbf{Negative}} & \textbf{0.1} & \textbf{0.2} & \textbf{0.3} & \textbf{0.4} & \textbf{0.5} & \textbf{0.6} & \textbf{0.7} & \textbf{0.8} & \textbf{0.9} & \textbf{1.0} & \textbf{Average} \\
    \hline
    \multirow{3}{*}{\makecell*[c]{\textbf{Perplexity($\downarrow$)}}} & \textbf{Linear} &{15.70} &{15.78} &{15.55} &{15.81} &{15.67} &{15.57} &{15.93} &{16.37} &{15.79} &{16.86} &{15.90} \\ 
    \cline{2-13}
    & \textbf{Union}
    &{14.22} &{14.32} &{14.17} &{14.23} &{14.73} &{14.57} &{14.63} &{14.45} &{14.78} &{14.90} &{14.50} \\
    \cline{2-13}
    &\textbf{Ours}
    &\textbf{6.46} &\textbf{6.42} &\textbf{6.49} &\textbf{6.46} &\textbf{6.60} &\textbf{6.71} &\textbf{6.68} &\textbf{6.62} &\textbf{6.78} &\textbf{6.63} &\textbf{6.59} \\
    \hline
    \multirow{3}{*}{\makecell*[c]{\textbf{Sentiment($\uparrow$)}}} & \textbf{Linear}  &{0.570} &{0.557} &{0.551} &{0.558} &{0.555} &{0.553} &{0.556} &{0.557} &{0.548} &{0.554} &{0.556} \\  
    \cline{2-13}
    & \textbf{Union}
    &{0.452} &{0.455} &{0.474} &{0.475} &{0.465} &{0.462} &{0.469} &{0.457} &{0.466} &{0.474} &{0.465} \\
    \cline{2-13}
    &\textbf{Ours}
    &\textbf{0.574} &\textbf{0.571} &\textbf{0.575} &\textbf{0.581} &\textbf{0.569} &\textbf{0.571} &\textbf{0.569} &\textbf{0.562} &\textbf{0.553} &\textbf{0.560} &\textbf{0.569}  \\
    \bottomrule
    \end{tabular}%
    }
    \caption{Multi-attributes on positive sentiment and negative sentiment control evaluation.}
  \label{tab:multi_over_result}%
\end{table*}%

\subsection{Multi-attribute control}
\label{sec:multi_control}
In this part, we utilize a dual-attribute control setting (toxicity \& sentiment) to evaluate the overlapping alleviation in the multi-attribute combination scenario. Concretely, we conduct experiment on dataset exact to that in Section \ref{sec:sent_control}, nevertheless, we write after the first 32 tokens for consistency without any sentiment transition.
We employ toxicity and sentiment as the control aspects, and manually add overlapping to bring conflict between them. Prompts for both the attributes are shown in Table \ref{tab:prompts_overlapping}. Taking \textbf{Setting one} for example, the \textbf{toxicity} attribute covers an extra antagonistic sentiment, i.e., \emph{slight negative}, to the \textbf{positive reply} attribute, which begets overlapping when combined in a linear manner. We select Llama2-7b as the basic language model. According to Table \ref{tab:multi_over_result}, where the coefficients from 0.1 to 1.0 mean the relative strength ratio, i.e., $s_{1}/s_{2}$, our strategy shows superiority in both Positive and Negative sentiment control settings that the average relative scores on sentiment are $+2.3\%$ and $+1.3\%$. As for the \textbf{Union} baseline, we set the sentiment attribute as the base, and vary the coefficient of $max(toxi, senti)$ from 0.1 to 1.0. In comparison with it, our method still dominates in both control settings. Hence it proves that the improved attribute combination can diminish the attribute conflict degree to a certain extent.

Moreover, we design another overlapping type with toxicity attribute covering a same sentiment trend in comparison with the sentiment attribute (details are shown in Section \ref{sec:overlapping_same}). Intuitively, we will derive a higher sentiment score in contrast to the previous design. In addition, the score growth of our method should fall short of the linear combination, as the consequence of that the enhanced overlapping should get deflated. Referring to Table \ref{tab:multi_over_result_same}, our method obtains increased sentiment scores of 0.189\% and 0.192\% by average on positive and negative sentiment settings separately, which are both less than the values, i.e., 0.577\% and 0.694\%, of the linear strategy.

Referring to $n>2$ attributes combination, we incorporate a new attribute that influences the sentiment reply as well on the basis of setting one in Table \ref{tab:prompts_overlapping}:

\newtcolorbox{box_A}{
    sharpish corners, 
    boxrule = 0pt,
    toprule = 4.5pt, 
    enhanced,
    fuzzy shadow = {0pt}{-2pt}{-0.5pt}{0.5pt}{black!35} 
}
\begin{box_A}
\textbf{Child's tone:} Writing after the movie review with mimicking the child's tone, and with some negative sentiment.
\end{box_A} 

\begin{table}[ht!]
    \small
  \centering
    \begin{tabular}{c|c|c|c|c|c}
    \toprule
    \textbf{} & \textbf{0.2}&\textbf{0.4}&\textbf{0.6}&\textbf{0.8}&\textbf{1.0}\\
    \hline
    \textbf{Linear} &{0.531} & {0.530} &{0.528} &{0.521} &{0.527}\\
    \hline
    \textbf{Ours} &\textbf{0.562} & \textbf{0.559} &\textbf{0.550} &\textbf{0.543} &\textbf{0.544}\\
    \bottomrule
    \end{tabular}%
    \caption{Attributes and prompts for sentiment transition.}
  \label{tab:3_attributes}%
\end{table}%
where the coefficients above means $s_{1}/s_{3} = s_{2}/s_{3}$ with also the \emph{sentiment control} as the reference.

\section{Conclusions}
Considering underlying overlapping between attributes, we deduce \textbf{Palette of Language Models}, an improved linear combination strategy for multi-attribute control, with \emph{the Law of Total Probability} and \emph{Conditional Mutual Information Minimization}. Different from previous linear combination methods, the derived formula owns a dynamic coefficient to each model which can enhance the attribute expression. Additionally, two pivotal properties are proposed that serve as guiding principles for designing a rational attribute combination strategy. We conduct comprehensive experiments on both single and multiple attributes control settings which further underscore the effectiveness of our method.


\section*{Limitations}
The \textbf{Palette of Language Models} we proposed has undergone relatively rigorous theoretical derivation and is suitable for autoregressive language models (with specific attributes) combination scenario. However, in actual applications, language models may be of different types, and the corresponding vocabulary may also vary. Therefore, we will solve the decoding problems caused by inconsistent vocabularies in the future work, trying to pave the way for more sophisticated and nuanced control mechanisms that can better cater to the extensive needs of language model applications.

\section*{Acknowledgements}
This work is supported by Beijing Natural Science Foundation
(L222006) and China Mobile Holistic Artificial Intelligence Major Project
Funding (R22105ZS, R22105ZSC01).


\bibliography{acl_latex}

\newpage

\appendix
\section{Details for Approximation Combination over Attribute Couple}
\label{sec:ac_detail}
We add single factorization on attributes $A_{i}$\&$A_{j}$ with the couple factorization on them, and get:
\begin{gather}
    3*p(Z)=\lambda_{i}*p(A_{i}=x) + \lambda_{i^{'}}*p(A_{i} \neq x) \notag \\ 
    + \lambda_{j}*p(A_{j}=x) + \lambda_{j^{'}}*p(A_{j} \neq x) \notag \\
   + \lambda_{ij}*p(A_{i}=x)p(A_{j}=x) \notag \\
    +\lambda_{i^{'}j^{'}}*p(A_{i} \neq x)p(A_{j}\neq x)\notag\\
+\lambda_{ij^{'}}*p(A_{i}=x)p(A_{j}\neq x) \notag\\
    +\lambda_{i^{'}j}*p(A_{i}\neq x)p(A_{j}=x)
    \label{equ:fact_3p}
\end{gather}
Furthermore, employing the convexity of the logarithmic function, i.e., $log(ax+by) - log(a+b)\geq \frac{a}{a+b}log(x)+\frac{b}{a+b}log(y)$, we approximate the equation above as:
\begin{gather}
    log3p(Z=x) \approx \lambda_{i}logp(A_{i}=x) + \lambda_{i^{'}}logp(A_{i} \neq x)\notag \\ 
    + \lambda_{j}logp(A_{j}=x) + \lambda_{j^{'}}logp(A_{j} \neq x) \notag \\
   + \lambda_{ij}log(p(A_{i}=x)p(A_{j}=x)) \notag \\
    +\lambda_{i^{'}j^{'}}log(p(A_{i} \neq x)p(A_{j}\neq x))\notag\\
    +\lambda_{ij^{'}}log(p(A_{i}=x)p(A_{j}\neq x)) \notag \\
    +\lambda_{i^{'}j}log(p(A_{i}\neq x)p(A_{j}=x))\notag \\
    = (\lambda_{i}+\lambda_{ij}+\lambda_{ij^{'}})logp(A_{i}=x)\notag \\+(\lambda_{i^{'}}+\lambda_{i^{'}j}+\lambda_{i^{'}j^{'}})logp(A_{i} \neq x) \notag \\
    + (\lambda_{j}+\lambda_{ij}+\lambda_{i^{'}j})logp(A_{j}=x)\notag\\+(\lambda_{j^{'}}+\lambda_{ij^{'}}+\lambda_{i^{'}j^{'}})logp(A_{j} \neq x)
    \label{equ:fact_3p_appr}
\end{gather}

\section{Proof for Conditional Independent between Attribute Couple}
\label{sec:pr_ci}
As for conditional mutual information $I(A_{i},A_{j}|Z)$, when over the token $x$ (which means $Z=x$), we will obtain:
\begin{gather}
    \sum_{a_{i},a_{j}} p(a_{i})p(a_{j})p(Z=x|a_{i},a_{j})*\notag\\
    log\frac{p(Z=x|a_{i},a_{j})p(Z=x)}{p(Z=x|a_{i})p(Z=x|a_{j})}
\end{gather}
Therefore, the simple operation on minimization is to satisfy the equation $p(Z=x|a_{i},a_{j})p(Z=x)=p(Z=x|a_{i})p(Z=x|a_{j})$.

\section{Equation \ref{equ:improve_linear} Simplification}
\label{sec:simp}
Based on \textbf{Cauchy inequality}, the inequalities establish that:
\begin{gather}
    2 \geq (\lambda_{j}+\lambda_{j^{'}}) \geq \sqrt{\lambda_{j}^{2}+\lambda_{j^{'}}^{2} }, \notag \\
    \sqrt{\lambda_{j}^{2}+\lambda_{j^{'}}^{2} }*\sqrt{p(A_{j}=x)^{2}+p(A_{j}\neq x)^{2} } \notag\\
    \geq (\lambda_{j}p(A_{j}=x) + \lambda_{j^{'}}p(A_{j}\neq x)) = p(Z=x)
\end{gather}
hence, we derive that the value of $p(A_{i}|Z=x)\sum_{j \neq i}(\lambda_{j}+\lambda_{j^{'}})$ is in the interval of $[(n-1)p(A_{i}=x|Z=x),2(n-1)p(A_{i}=x|Z=x)]$, namely that $(n-1)\lambda_{i}$ and $p(A_{i}|Z=x)\sum_{j \neq i}(\lambda_{j}+\lambda_{j^{'}})$ are in the same order of magnitude. Therefore, we approximate $p(A_{i}|Z=x)\sum_{j \neq i}(\lambda_{j}+\lambda_{j^{'}})$ with $(n-1)\lambda_{i}$ for simplification.

Furthermore, we simplify $\varphi_{i}$ and $\omega_{i}$ with highlighting the kernel part $1+\frac{1}{p(A_{i})}$ and condensing other variables as $s_{i}$ to express the strength (the proportion to the final generation) of the current attribute. We consider $logp(A_{i} \neq x)$ part as an auxiliary to $logp(A_{i}=x)$, thus introduce a small coefficient $t$ ahead.

\section{Proof for Strategy Properties}
\label{sec:psp}
\paragraph{Proof for property 1} 
According to Equation \ref{equ:comb_norm}, we focus on the main part (i.e. $p(A_{i}=x)$) and obtain that:
\begin{gather}
p(Z\!\!=\!\!x_{k})\!\!=\!\!\frac{Q_{k}*p(A_{i}=x_{k})^{s_{i}*(1+\frac{1}{p(A_{i}=x_{k})})}}{\sum_{v \in V} Q_{v}*p(A_{i}=x_{v})^{s_{i}*(1+\frac{1}{p(A_{i}=x_{v})})}}\notag \\
    = \frac{1}{\sum_{v \in V} \frac{Q_{v}}{Q_{k}} * p_{vk}^{s_{i}}} \notag \\
    \textbf{s.t.} \hspace{4pt} Q_{k}=P_{b}(x_{k})\prod_{j \neq i} p(A_{j}=x_{k})^{s_{j}*(1+\frac{1}{p(A_{j}=x_{k})})}, \notag\\
    p_{vk} = \frac{p(A_{i}=x_{v})^{1+\frac{1}{p(A_{i}=x_{v})}}}{p(A_{i}=x_{k})^{1+\frac{1}{p(A_{i}=x_{k})}}}
    \label{equ:pos_corr}
\end{gather}
where $x_{k}$ is the attribute token on $A_{i}$. Considering \emph{attribute token} definition in Section \ref{sec:propery}, we conclude $p_{vk}$ is less than 1 (details are shown in Section \ref{sec:pvb}), therefore, when attribute strength $s_{i}$ increases ($Q_{k}$ is fixed), the final generation probability will grow subsequently.
\paragraph{Proof for property 2} 
Similarly, we also consider the main part in Equation \ref{equ:comb_norm}. Taking into account the gaps between attribute token $x_{attr}$ and non-attribute token $x_{v}$ of both methods, we get:
\begin{gather}
    (\textbf{ours}) \hspace{4pt} gap \!\!=\!\! s_{i}[(1\!\!+\!\!\frac{1}{p(A_{i}\!\!=\!\!x_{attr})})logp(A_{i}=x_{attr})\notag\\
    - (1\!\!+\!\!\frac{1}{p(A_{i}\!\!=\!\!x_{v})})logp(A_{i}\!\!=\!\!x_{v})]\notag \\
    (\textbf{linear}) \hspace{4pt} gap = s_{i}(logp(A_{i}=x_{attr})\notag\\
    -logp(A_{i}=x_{v}))
    \label{equ:our_enh}
\end{gather}
Referring to a special function $f(x)=\frac{logx}{x} - logx$ which is always increasing within the interval of $(0,1)$, we derive that the gap of our strategy is greater than that of linear combination. Thus, if distributions on other attributes are fixed, the final generation tends to perform more like current attribute in our strategy.

\section{Proof for $p_{vk}<1$}
\label{sec:pvb}
\begin{gather}
p(A_{i}=x_{k})^{1+\frac{1}{p(A_{i}=x_{k})}} > 
p(A_{i}=x_{k})^{1+\frac{1}{p(A_{i}=x_{v})}} \notag \\ >
p(A_{i}=x_{v})^{1+\frac{1}{p(A_{i}=x_{v})}} \notag \\
\textbf{s.t.} \hspace{4pt} p(A_{i}=x_{k}) > p(A_{i}=x_{v})
\end{gather}
Hence the inequality $p_{vk}<1$ is satisfied.

\section{Proof for rationality of $\sigma(p(A_{i}=x))$}
\label{sec:sigma_proof}
The surrogate stuff of $p(A_{i}=x)$ on the denominator (in Equation \ref{equ:comb_norm}) is rational when it satisfies both the properties of \textbf{Theorem 1} and \textbf{Theorem 2}.
\paragraph{Proof for property 1} Similar to Equation \ref{equ:pos_corr}, we substitute $\sigma(p(A_{i})=x)$ for $p(A_{i}=x)$ on the denominator and obtain:
\begin{gather}
    p(Z\!\!=\!\!x_{k})\!\!=\!\!\frac{Q_{k}*p(A_{i}=x_{k})^{s_{i}*(2+e^{-p(A_{i}\!\!=\!\!x_{k})})}}{\sum_{v \in V} Q_{v}*p(A_{i}=x_{v})^{s_{i}*(2+e^{-p(A_{i}\!\!=\!\!x_{v})})}}
    \notag \\
    = \frac{1}{\sum_{v \in V} \frac{Q_{v}}{Q_{k}} * p_{vk}^{s_{i}}} \notag \\
    \textbf{s.t.} \hspace{4pt} Q_{k}=P_{b}(x_{k})\prod_{j \neq i} p(A_{j}=x_{k})^{s_{j}*(2+e^{-p(A_{j}=x_{k})})}, \notag\\
    p_{vk} = \frac{p(A_{i}=x_{v})^{2+e^{-p(A_{i}=x_{v})}}}{p(A_{i}=x_{k})^{2+e^{-p(A_{i}=x_{k})}}}
    \label{equ:pos_corr_sigma}
\end{gather}
Likewise, $x_k$ is the \emph{Attribute Token} which satisfies $p(A_{i}=x_{k})>p(A_{i}=x_{v})$. Furthermore, the inequality $e^{-p(A_{i}=x_{k})}<e^{-p(A_{i}=x_{v})}$ will hold. Hence, we will also obtain that $p_{vk}<1$ and there exists a positive correlation between $s_{i}$ and $p(Z=x_{k})$.

\paragraph{Proof for property 2} Following Equation \ref{equ:att_enh}, we tick off gaps of both the $\sigma$ substitute and the linear combination, as:
\begin{gather}
    (\textbf{$\sigma$}) \hspace{4pt} gap = s_{i}[(2+e^{-p(A_{i}=x_{attr})})logp(A_{i}=x_{attr}) \notag\\
    - (2+e^{-p(A_{i}=x_{v})})logp(A_{i}=x_{v})]\notag \\
    (\textbf{linear}) \hspace{4pt} gap = s_{i}(logp(A_{i}=x_{attr}) \notag\\
    -logp(A_{i}=x_{v}))
    \label{equ:our_enh_sigma}
\end{gather}
We introduce a special function and its derivative:
\begin{gather}
    f(x) = (2+e^{-x})logx - logx,\notag\\
    \nabla f(x) = \frac{1+e^{-x}-xe^{-x}logx}{x}
\end{gather}
Obviously, $f(x)$ is increasing in the interval of $(0,1)$ due to $\nabla f(x)$ being always greater than 0. Therefore, we derive that $(2+e^{-p(A_{i}=x_{attr})})logp(A_{i}=x_{attr})-logp(A_{i}=x_{attr})$ is greater than $(2+e^{-p(A_{i}=x_{v})})logp(A_{i}=x_{v})-logp(A_{i}=x_{v})$ which means the \textbf{gap} of the $\sigma$ substitute outweighs that of the linear combination.

\section{Results for complementary event evaluation}
\label{sec:ce_pic}
\begin{figure*}[h]
\centering
\includegraphics[width=\textwidth]{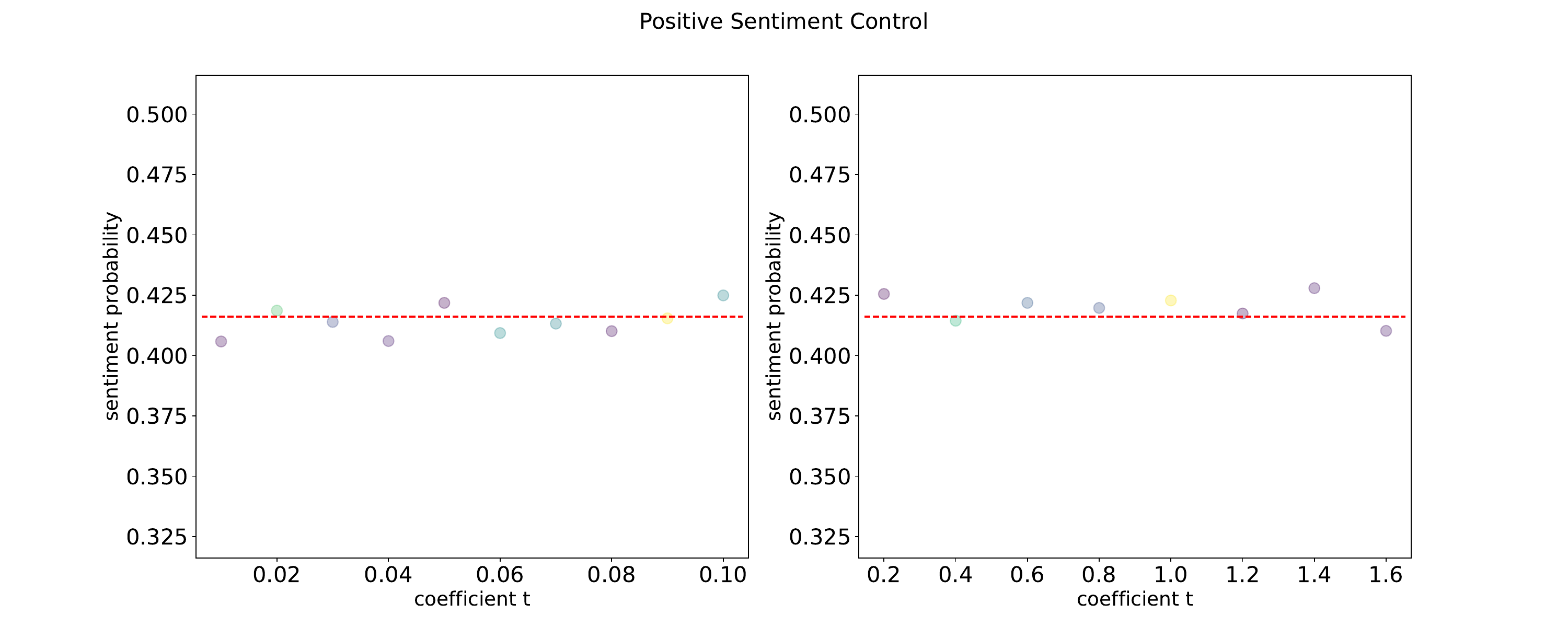}
\caption{Coefficient $t$ of $logp(A_{i} \neq x)$ evaluation on Positive Sentiment Control scenario.}
\label{fig:pos_t}
\end{figure*}

\begin{figure*}[h]
\centering
\includegraphics[width=\textwidth]{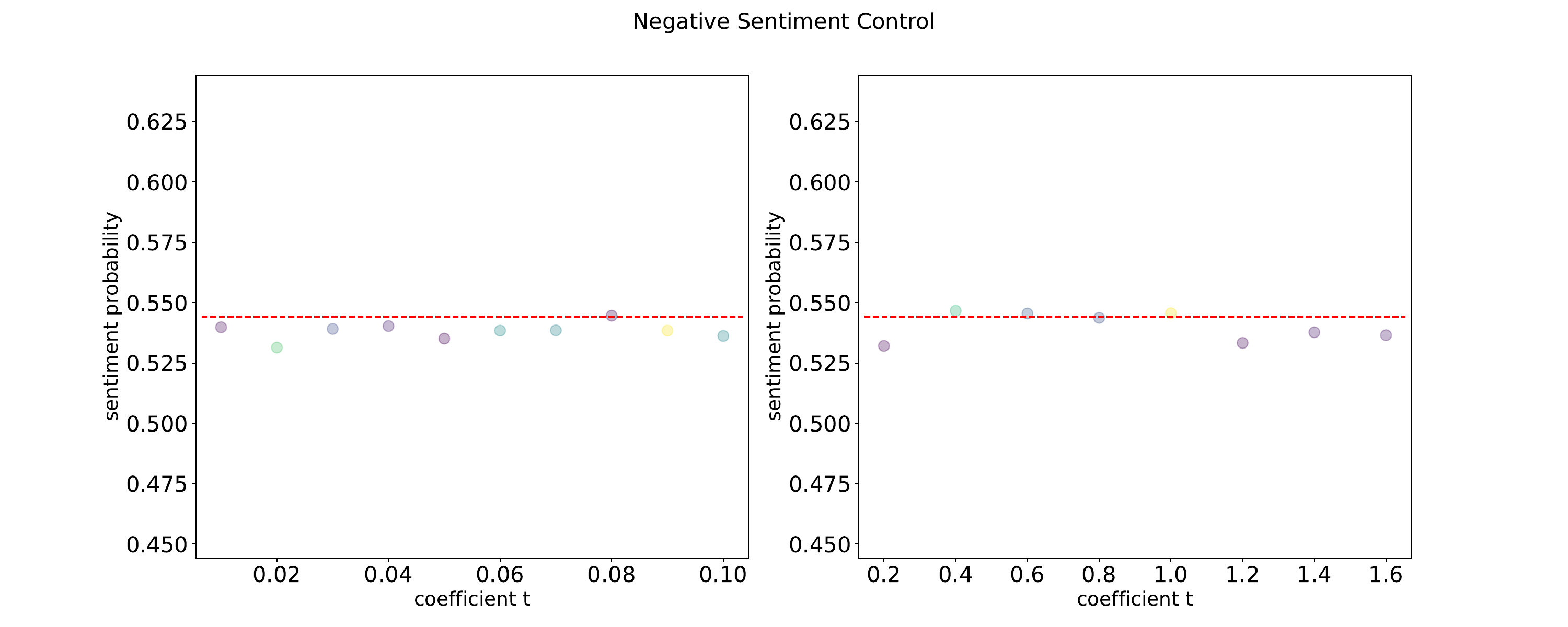}
\caption{Coefficient $t$ of $logp(A_{i} \neq x)$ evaluation on Negative Sentiment Control scenario.}
\label{fig:neg_t}
\end{figure*}

We conduct experiments on the verification of $t$ in Equation \ref{equ:comb_norm} and the results are demonstrated in Figures \ref{fig:pos_t} \& \ref{fig:neg_t}.

\section{The same trend overlapping setting for multiple attributes combination}
\label{sec:overlapping_same}

As is shown in Table \ref{tab:prompts_overlapping_same}, both the settings are designed under the same trend overlapping condition, namely that attribute one will enhance the performance of attribute two. 

We calculate the differential scores between overlapping settings of the same trend and the converse trend (Table \ref{tab:multi_over_result}), and report the results in Table \ref{tab:multi_over_result_same}.
\begin{table*}[ht!]
  \centering
    \begin{tabular}{ccc}
    \toprule
    \textbf{Settings} & \textbf{Attribute} & \textbf{Prompt}\\
    \hline
    \multirow{2}{*}{\textbf{\makecell[c]{Setting \\one}}} & 
    \textbf{toxicity} & \makecell[c]{The following movie review is not threatening and does not contain any\\sexually explicit or profane language, and is with a slight \textbf{positive} sentiment.} \\ 
    \cline{2-3}
    & \textbf{\makecell[c]{positive \\reply}} & \makecell[c]{Continue writing after the movie review to change it with \\ a very \textbf{positive} sentiment and a very \textbf{positive} tone.}\\
    \hline
    \multirow{2}{*}{{\textbf{\makecell[c]{Setting \\two}}}} & 
    \textbf{toxicity} & \makecell[c]{The following movie review is not threatening and does not contain any\\sexually explicit or profane language, and is with a slight \textbf{negative} sentiment.} \\ 
    \cline{2-3}
    & \textbf{\makecell[c]{negative \\reply}} & \makecell[c]{Continue writing after the movie review to change it with \\a very \textbf{negative} sentiment and a very \textbf{negative} tone.}\\
    \bottomrule
    \end{tabular}%
    \caption{Attributes and prompts for the multi-control with the same sentiment trend overlapping setting.}
  \label{tab:prompts_overlapping_same}%
\end{table*}%

\begin{table*}[h!]
  \centering
  \resizebox{\linewidth}{!}{
    \begin{tabular}{ccccccccccccc}
    \toprule
    \textbf{Positive} & \textbf{0.1} & \textbf{0.2} & \textbf{0.3} & \textbf{0.4} & \textbf{0.5} & \textbf{0.6} & \textbf{0.7} & \textbf{0.8} & \textbf{0.9} & \textbf{1.0} & \textbf{Average} \\
    \hline
    \textbf{Linear}&{0.020} &{0.990} &{0.530} &{1.010} &{1.030} &{-0.760} &{0.670} &{0.710} &{0.330} &{1.240} &{0.577}  \\
    \textbf{Ours}
    &{-0.430} &{0.590} &{0.240} &{1.180} &{-0.050} &{0.190} &{0.170} &{0.180} &{-0.670} &{0.490} &\textbf{0.189} \\
    \Xhline{1.5pt}
    \textbf{Negative} & \textbf{0.1} & \textbf{0.2} & \textbf{0.3} & \textbf{0.4} & \textbf{0.5} & \textbf{0.6} & \textbf{0.7} & \textbf{0.8} & \textbf{0.9} & \textbf{1.0} & \textbf{Average} \\
    \hline
    \textbf{Linear}
    &{-0.270} &{0.800} &{1.220} &{0.540} &{0.610} &{1.640} &{0.500} &{0.130} &{1.030} &{0.740} &{0.694} \\
    \textbf{Ours}&{0.140} &{0.680} &{-0.280} &{-1.910} &{0.500} &{-0.370} &{0.090} &{0.400} &{1.420} &{1.250} &\textbf{0.192}  \\
    \bottomrule
    \end{tabular}%
    }
    \caption{Sentiment Increase (\%) on the basis of Table \ref{tab:multi_over_result}, less is better.}
  \label{tab:multi_over_result_same}%
\end{table*}%

\end{document}